\documentclass[11pt]{article} % For LaTeX2e
\usepackage{rldmsubmit,palatino}
\usepackage{graphicx}
\usepackage[round]{natbib}
\usepackage{wrapfig}
\usepackage{subfig}

\DeclareUnicodeCharacter{0301}{*************************************}
\DeclareUnicodeCharacter{0307}{*************************************}

\title{Characterizing the Action-Generalization Gap in Deep Q-Learning}

\author{
Zhiyuan Zhou \\
Department of Computer Science\\
Brown University\\
Providence, RI 02912 \\
\texttt{zhouzy@brown.edu} \\
\And
Cameron Allen \\
Department of Computer Science \\
Brown University \\
Providence, RI 02912 \\
\texttt{csal@brown.edu} \\
\AND
Kavosh Asadi \\
Amazon Web Services \\
Santa Clara, CA 95054 \\
\texttt{kavasadi@amazon.com} \\
\hphantom{Department of Computer Science} \\ % this is to make the spacing line up correctly
\And
George Konidaris \\
Department of Computer Science \\
Brown University \\
Providence, RI 02912 \\
\texttt{gdk@cs.brown.edu} \\
}

\begin{document}

\maketitle

\begin{abstract}
We study the action generalization ability of deep Q-learning in discrete action spaces. Generalization is crucial for efficient reinforcement learning (RL) because it allows agents to use knowledge learned from past experiences on new tasks.
% But while deep RL agents have a natural way of generalizing over states through function approximation, one might think agents cannot utilize the same mechanism in discrete action domains because actions are outputs of the function approximator.
But while function approximation provides deep RL agents with a natural way to generalize over state inputs, the same generalization mechanism does not apply to discrete action outputs.
And yet, surprisingly, our experiments indicate that Deep Q-Networks (DQN), which use exactly this type of function approximator, are still able to achieve modest action generalization. Our main contribution is twofold: first, we propose a method of evaluating action generalization using expert knowledge of action similarity, and empirically confirm that action generalization leads to faster learning; second, we characterize the action-generalization gap (the difference in learning performance between DQN and the expert) in different domains. We find that DQN can indeed generalize over actions in several simple domains, but that its ability to do so decreases as the action space grows larger.

\end{abstract}


\acknowledgements{This research was supported by a Brown University Undergraduate Teaching and Research Award, as well as by the ONR under the PERISCOPE MURI Contract N00014-17-1-2699, and by the NSF under grant 1955361.}

\startmain % to start the main 1-4 pages of the submission.

\section*{Introduction}
In reinforcement learning, generalization~\citep{ponsen2009abstraction} is
% the agent's ability to extrapolate information from environment data they have observed and apply it to adapt to new unseen situations. It is
crucial for achieving efficient learning and leads to better performance over unseen data.
Generalization allows agents to extrapolate from environment data they have observed and to adapt to unseen situations.
It is well known that deep learning supports generalization~\citep{dl_generalization}, and deep reinforcement learning agents, 
% a natural way of generalizing over inputs.
which use deep learning to maximize reward, can therefore be expected to generalize as well.
However, this kind of generalization is assumed to come from parameter sharing in the function approximator, and thus only provides a natural way of generalizing over inputs.
% zzz: refs for generalization in DRL?
% zzz: refs for generalization through param sharing?

In discrete-action domains, deep reinforcement learning (DRL) agents typically only use states as inputs, because incorporating actions as inputs becomes computationally expensive as the size of the action space grows. Therefore, such agents, typified by DQN~\citep{mnih2015human}, cannot rely on the same parameter sharing mechanism for generalizing over actions. 
And yet, many of the domains on which DQN has been shown to perform well have action spaces where generalization is not only possible, but essential. In particular, many of the Atari 2600 games~\citep{machado2018revisiting} include multiple actions with identical effects or actions (e.g. RIGHTFIRE) that combine the effects of two or more other actions (e.g. RIGHT and FIRE). DQN's ability to cope with such action spaces is therefore surprising, given that it is unclear exactly how it is generalizing over these kinds of ``similar'' actions.
% zzz: ref for computationally intractability, cite such DQN agents?
% But it is essential that agents are able to do action generalization if they were to learn about an unseen action based on actions already seen.
% It is therefore important to determine the extent to which existing DRL agents can generalize over actions.
% We choose to examine the DQN agent because it is arguably the most well-understood deep Q-learning algorithm with widespread influence.

In this work, we seek to better understand DQN's (arguably counter-intuitive) ability to generalize over actions. We first empirically confirm the widely-held belief that action generalization leads to faster learning. We introduce an oracle for characterizing ``perfect'' action generalization with DQN using expert knowledge: when it is known which actions lead to the same transition effects, all action values for that subset are updated in one Q-update.
% This represents the performance ceiling of an agent whose actions are fully generalized.
Our experiments indicate that such generalization indeed improves learning speed. Next, we study action generalization in unmodified DQN by measuring its learning performance against the oracle; the difference is what we term the \emph{action-generalization gap}. We experiment on classic Gym control environments~\citep{brockman2016openai} and Atari 2600 games, and find that DQN's ability to generalize over actions depends on the size of the action space. In small action spaces, DQN performs nearly as well as the expert; however, when the action space gets large, DQN performs poorly because of its inability to generalize.

\section*{Expert Action Generalization}
One way of achieving action generalization is through action abstraction. In the context of reinforcement learning, abstraction is a method to map the representation of the original problem to a new, simpler representation where irrelevant properties are filtered and only properties relevant to decision-making are kept~\citep{abel-thesis, ponsen2009abstraction}. Action abstraction, in particular, is the technique of grouping similar actions together into one abstract action, ignoring their small differences that are not relevant to decision-making. It facilitates generalization by abstracting similar experiences, and allowing information about one experience to apply to related (unseen) experiences.

We introduce an oracle for characterizing perfect action generalization that uses expert knowledge to abstract over actions that are similar to each other. More precisely, in a Markov Decision Process with state space $S$, action space $A$, reward function $R$, and discount factor $\gamma$, the expert knowledge is a symmetric $|A|\times|A|$ similarity matrix $K$, where each index $K(i, j)$ is a value between $0$ and $1$ indicating the similarity score of actions $a_i$ and $a_j$. A score of $1$ means two actions are fully similar, and $0$ means fully different. Given this expert knowledge, we can adjust the Q-update process during Q-learning: for an experience tuple $(s, a, r, s')$, we update
$$ Q(s, \tilde{a}) = Q(s, \tilde{a}) + \alpha * K(a, \tilde{a}) * [(r + \gamma V(s')) - Q(s, \tilde{a})] \quad \forall\ \tilde{a} \in A.$$
In words, during each update step, we not only update $Q(s, a)$ from the experienced action $a$, but also $Q(s, \tilde{a})$, proportionally to how similar $a$ and $\tilde{a}$ are. So, a fully similar action $\tilde{a}$ will get a full Q update, and a fully dissimilar action $\tilde{a}$ will not be updated at all, and those in between will be updated proportionally. This process allows generalization to $\tilde{a}$ despite only experiencing $a$ in the environment. 

This oracle provides a way of measuring the degree of action generalization through learning performance. To see how much action generalization DQN can do, we can simply compare it with the oracle, which represents a best-case performance ceiling for action generalization methods. Since it is hard to quantify action generalization directly, we compare learning performance instead, using it as a proxy for generalization. We define this performance difference between the oracle and a DRL agent to be the \emph{action-generalization gap}. The rest of this paper aims to characterize the action-generalization gap for DQN.

\section*{Action Space Augmentation for Evaluation}
In order to evaluate action generalization, we need environments with action spaces where actions may be similar to each other, so that it makes sense for actions to generalize. To that end, we propose to augment the action space of a base environment to include additional similar actions. We propose three such ``action augmentation'' methods:
\begin{enumerate}
    \item \emph{Duplicate actions ($N\times$)}: augment the original action set $N-1$ times, so the new action space contains $N$ copies of every action in the original action space.
    \item \emph{Semi-duplicate actions}: augment the original action set with $4$ sets of reduced-magnitude actions, $|A|$ in each set (where $|A|$ is the size of the original action space), for a total of $5\cdot|A|$ actions. Each set of these semi-duplicate actions has similar transition effects to the original action set, differing only in the magnitude. The magnitude similarity is controlled by a similarity score $h \in [0, 1]$, where $h=1$ corresponds to the same-magnitude action and $h=0$ corresponds to a zero-magnitude action (No-op). For example, in the Pendulum environment, we can apply a $0.8$ torque to the left, or $0.5$ torque to the right, etc.
    \item \emph{Random actions}: augment the original action set with $4\cdot|A|$ stochastic actions for a total of $5\cdot|A|$ actions. Each stochastic action is a uniform random distribution over the actions in the original action space.
\end{enumerate}

Here we are concerned with discrete action spaces, so we take discrete-action versions of CartPole, Pendulum, and LunarLander as the base environments for our experiments, and apply the three action augmentation methods to expand the action space.

Because the action-augmentations above are artificially created, we can supply the oracle with knowledge of which similar actions should be abstracted together. For the ``duplicate actions'' augmentation, all duplicate copies of each actions are fully similar ($K(i, j) =1$) to each other, and not to anything else. For the ``semi-duplicate actions'' augmentation, $K(i, j) = h$ if $a_i$ and $a_j$ are semi-duplicates of each other, else $K(i, j) = 0$. For the ``random actions'' augmentation, all the random actions are fully similar to each other, and nothing else is similar. For all augmentations, $K(i, i) = 1$. The oracle uses this information to guide its Q updates.

\section*{Evaluating Performance}
We now experimentally characterize the action-generalization gap between DQN and the oracle. In addition to the augmented action space, we provide the original action space as a baseline for comparison, which we call ``baseline''. For the ``duplicate actions'' augmentation in this section, we set $N=5$. For the semi-duplicate augmentation, we use similarity scores $h \in \{0.2, 0.5, 0.8\}$.

\begin{figure}[!htbp]
    \centering
    \includegraphics[width=0.9\linewidth]{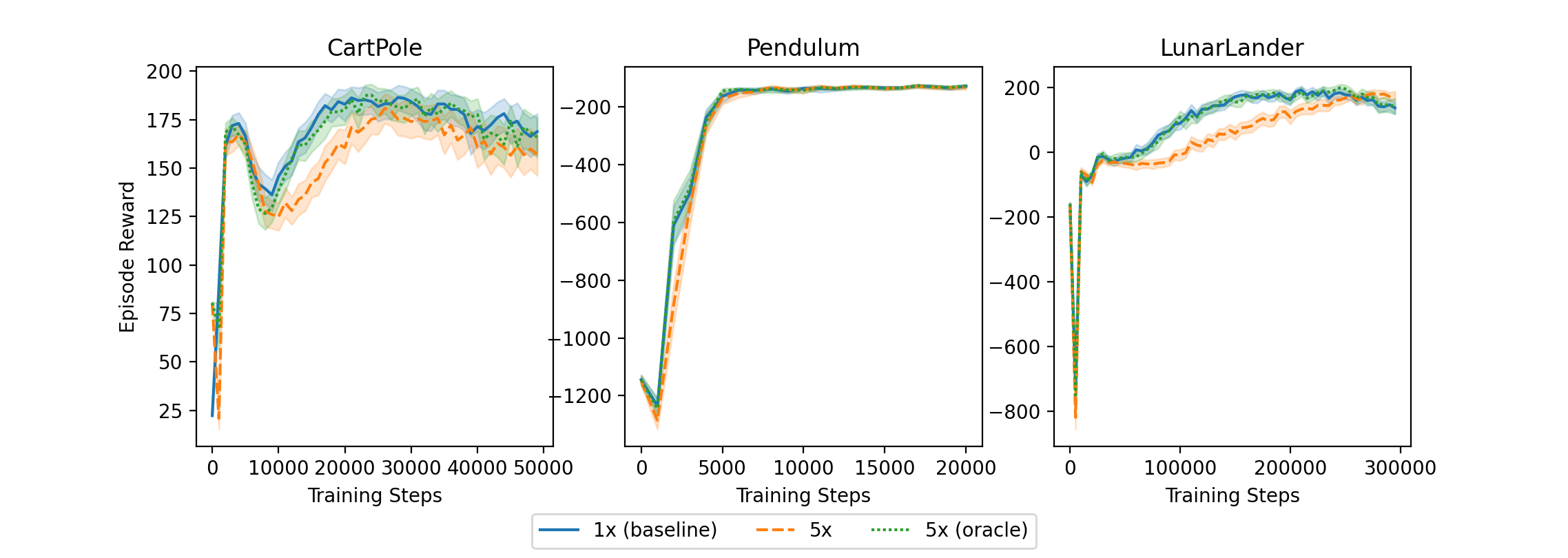}
    \caption{Action-generalization gap for DQN with \textbf{5x duplicate actions} on CartPole, Pendulum, and LunarLander}
    \label{fig:duplicate}
\end{figure}

In the ``duplicate actions'' environments, having duplicate actions slows down learning, with the exception of Pendulum, where 5x duplicate actions performs about the same as the baseline (Figure~\ref{fig:duplicate}). By contrast, the oracle is unaffected by the larger action space and performs just as well as the baseline.
% When the actions is fully generalized with the help of the oracle, the DQN learning performance on an augmented envionment is about the same as the unaugmented environment.
This confirms the hypothesis that action generalization does help speed up learning. 

The experiment also provides evidence that DQN is doing some implicit action generalization. Notice the difference in performance between the oracle and unmodified DQN in Figure~\ref{fig:duplicate}: DQN takes at most twice as long as the oracle to learn with 5x duplicate actions. This is a bit surprising given that the oracle performs the same as the unaugmented environment (baseline), whose action space is one-fifth as large. If DQN isn't doing any action generalization, then learning time should scale linearly with the size of the action space. The fact that it scales sub-linearly suggests that DQN is able to generalize over actions to some degree.

These 5x duplicate action results are somewhat surprising, because even though there are more actions to choose from, the probability of choosing the optimal action under a uniform random policy remains the same. So why does the learning problem become harder? We hypothesize two potential explanations. First, it could be a result of the overestimation bias in Q-Learning~\citep{thrun1993issues}, which increases with the size of the action space. Second, it could be that a larger action space means greater opportunity for noise in neural network parameters to produce noisy gradient updates. We leave a full investigation of these hypotheses for future work.

\begin{figure}[tbp!]
  \centering
  \hfill
  \begin{minipage}[c]{0.28\textwidth}
    \includegraphics[width=\textwidth]{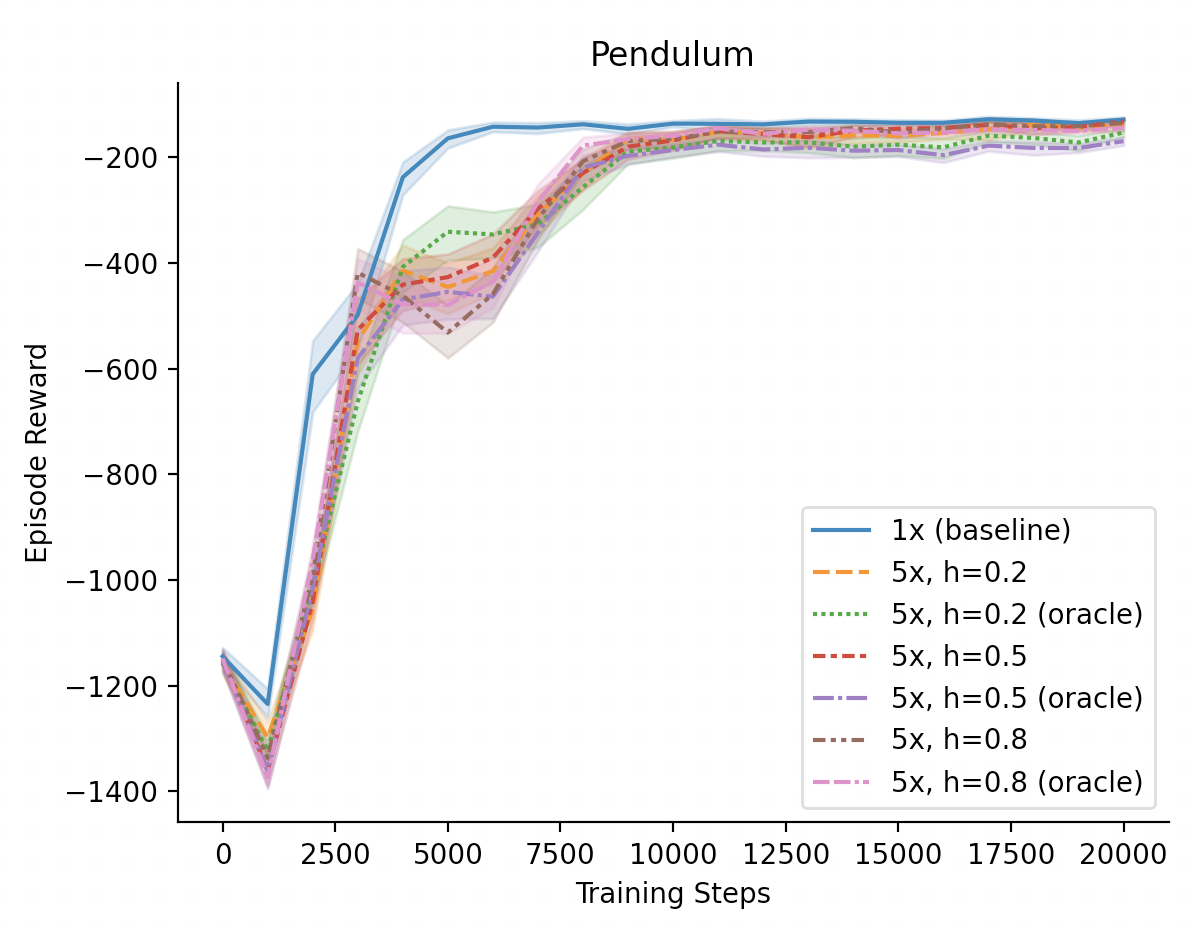}
    % \caption{.}
    % \label{fig:semi-duplicate}
  \end{minipage}
  \hfill
  \begin{minipage}[c]{0.68\textwidth}
    \includegraphics[width=\textwidth]{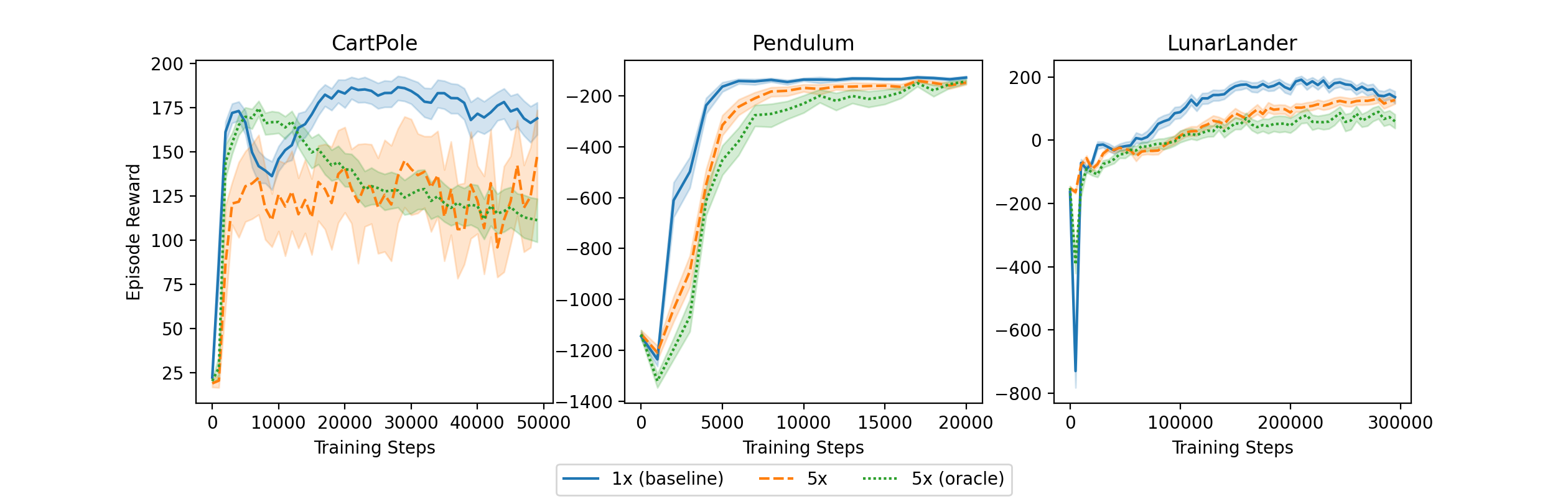}
    % \caption{.}
    % \label{fig:random}
  \end{minipage}
  \caption{(Left) Action-generalization gap for DQN with \textbf{5x semi-duplicate actions}, for similarity score $h \in \{0.2, 0.5, 0.8\}$, on Pendulum. (Right) Action-generalization gap for DQN with \textbf{5x random actions} on CartPole, Pendulum, and LunarLander.}
  \label{fig:semi-and-random}
  \hfill

\end{figure}

For the semi-duplicate action augmentations, all the experiments take about twice as long to converge, even the ones using an oracle (Figure~\ref{fig:semi-and-random}, left). This indicates that semi-duplicate actions form a harder learning problem than exactly-duplicated actions, as expected. But still, we can note here that there is basically no action-generalization gap. This is more evidence to suggest that DQN is able to generalize over actions in small action spaces.

The ``random actions'' (Figure~\ref{fig:semi-and-random}, right) results tell a more complicated story: when random actions are introduced, the learning performance is worse than ``baseline'', regardless of whether the oracle is used. This shows that having random actions makes learning substantially harder, which is expected because this stochasticity is hard to account for during learning. What's surprising is that in Pendulum and LunarLander the action-generalization gap is actually negative: the oracle performs worse than unmodified DQN. We suspect this may be due to an incorrect assumption: the oracle treats all random actions as similar to each other. However, the random actions are not always similar; in fact, they frequently select from original actions that have completely opposite effects. Performing Q-updates with the assumption that all random actions are fully similar may at times lead the function approximator to incorrectly update shared internal parameters governing the Q-values of the \emph{non-stochastic} actions. If we instead assume the random actions are fully dissimilar, we simply recover the baseline performance. So, in the case of random actions, ``baseline'' is a better performance ceiling because none of its ``abstract'' actions contains actions with opposite effects. Using this ceiling, the action-generalization gap is quite noticeable, indicating that DQN is not good at generalizing over stochastic actions.

\section*{Evaluation on Large Action Spaces}
The experiments from the last section suggest that DQN has some robustness to action augmentation in domains with small action spaces. Here we investigate whether that robustness extends to environments with larger action spaces, and find that, for both Pendulum and Atari 2600 games, it does not.

\begin{figure}[tbp!]
  \centering
  \hfill
  \begin{minipage}[c]{0.30\textwidth}
    \includegraphics[width=\textwidth]{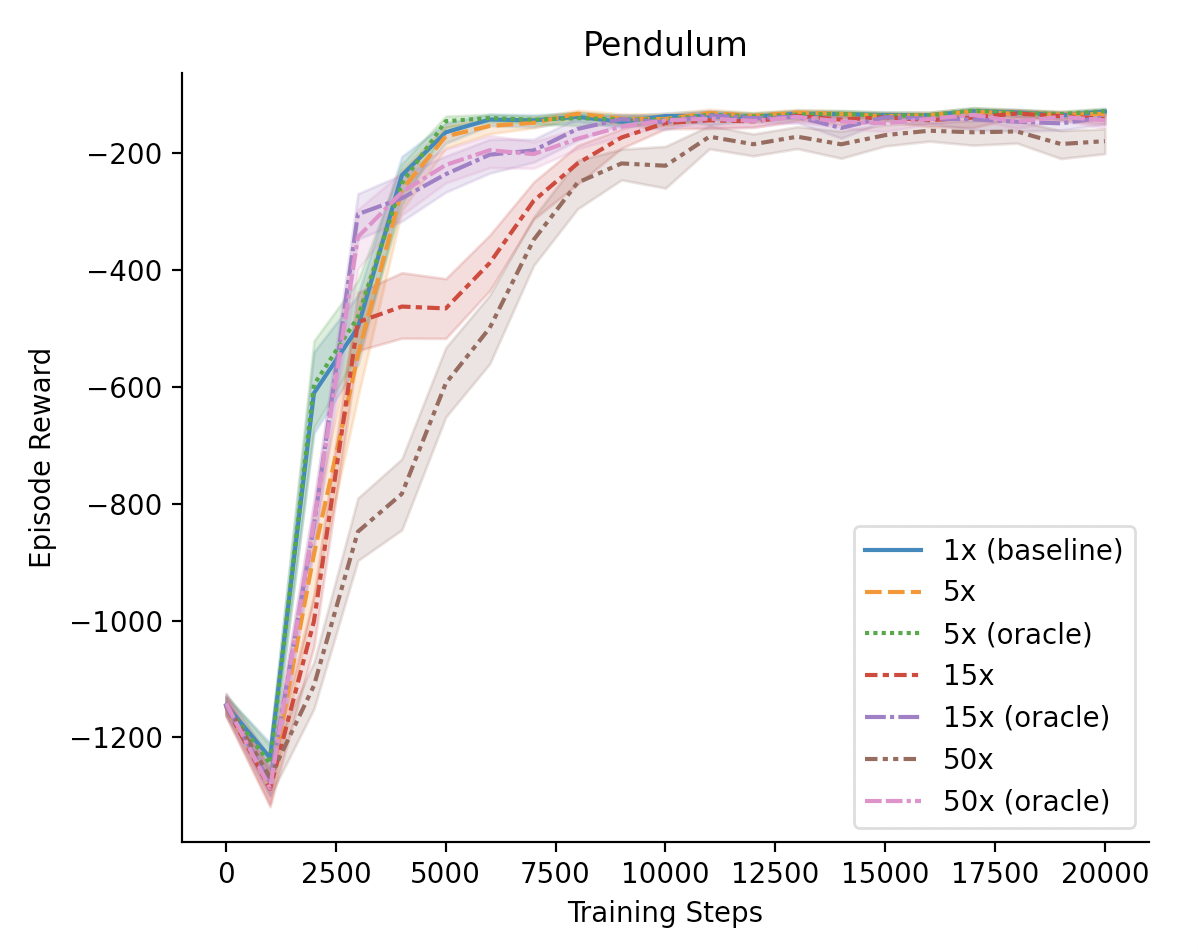}
    % \caption{.}
    % \label{fig:ndup-increasing}
  \end{minipage}
  \hfill
  \begin{minipage}[c]{0.68\textwidth}
    \includegraphics[width=\textwidth]{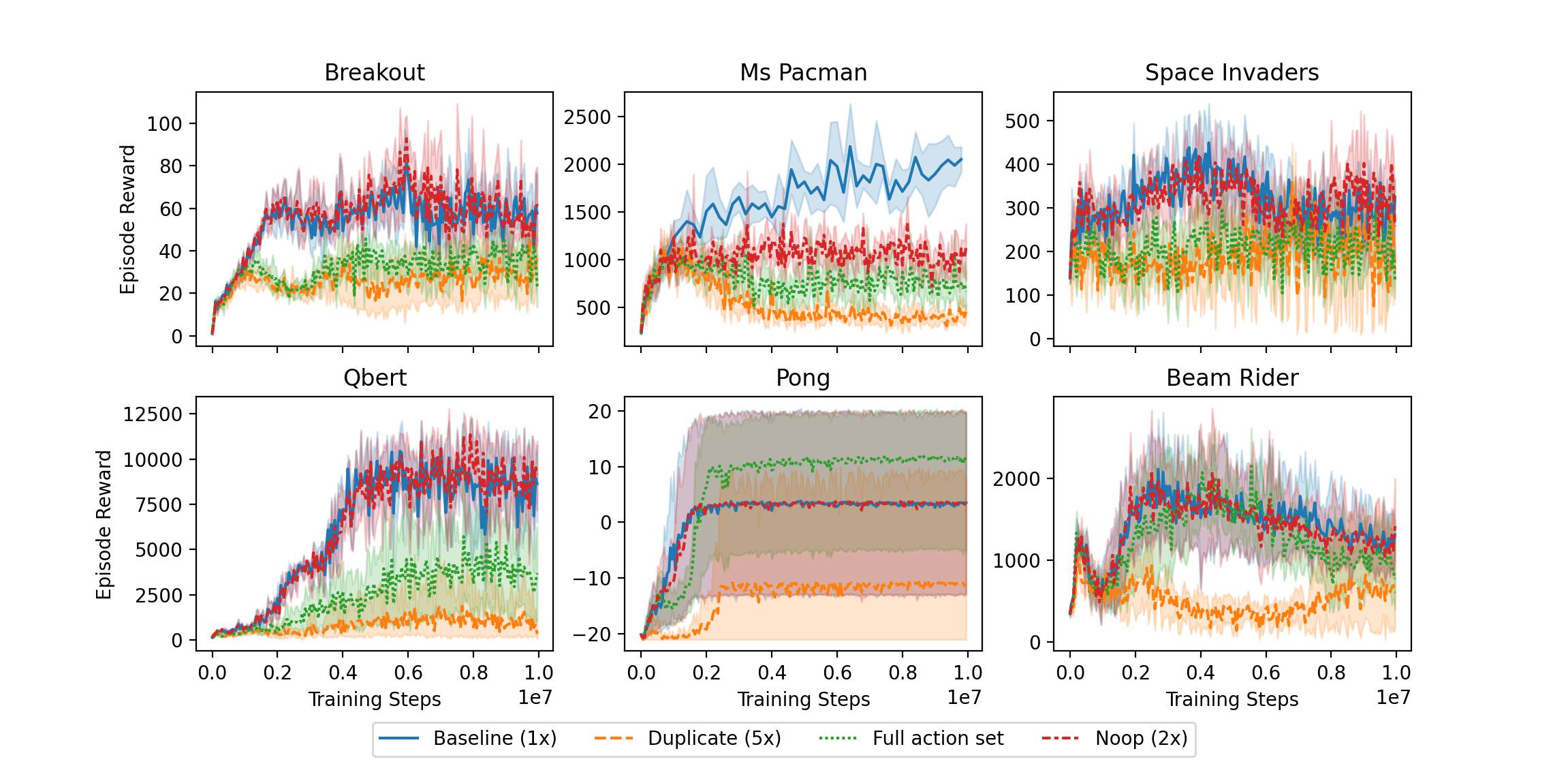}
    % \caption{.}
    % \label{fig:atari}
  \end{minipage}
  \caption{(Left) Action-generalization gap for DQN with \textbf{5x, 15x, and 50x duplicate actions} on Pendulum. (Right) Action-generalization gap for DQN on six Atari games.}
  \label{fig:large-scale}
  \hfill

\end{figure}

For Pendulum, we again augment with duplicate actions, and increase $N$ to discover the point at which learning performance starts to degrade (see 
% , and Atari 2600 games.
Figure~\ref{fig:large-scale}, left).
% shows the learning curve for ``N dup'' augmentations with increasing $N$.
When $N=5$, we do not observe any action-generalization gap, but when $N \in \{15, 50\}$, the gap becomes quite noticeable. Moreover, when $N=50$, learning is not just slow, but converges to worse final performance. This deterioration in DQN's performance is not due to the inherent hardness of the domain, because the oracle continues to match the performance of learning on the unaugmented environment. This suggests that the large action-generalization gap must come from DQN's inability to generalize over actions is large action spaces.

For Atari 2600, we chose six commonly-used games to evaluate the action-generalization gap: Beam Rider, Breakout, Pong, Ms Pacman, Qbert, and Space Invaders. Our experiments use three types of action augmentation:
\begin{enumerate}
    \item \emph{Full action set}: There are 18 legal actions that are shared across all Atari games. However, since some of the actions are not meaningful in certain games, the default action space of each game is usually pruned to only leave the meaningful action space (which we refer to as the ``baseline'' actions). In the full action set setting, we use all of the 18 legal actions in Atari games as the action space, which increases the size of the action space for all the games we investigate.
    \item \emph{Duplicate actions}: augment the baseline actions with $N-1$ sets of duplicate actions. Here we use $N=5$.
    \item \emph{Noop actions}: augment the baseline actions with $N\cdot|A|$ noop actions. Here we use $N=2$.
\end{enumerate}

In Atari games, we don't have expert knowledge of which actions are similar, because such information is game-specific and often highly state-dependent. Fortunately, we can still use the unmodified baseline action space as a performance ceiling, similar to the oracle in the preceding experiments. Even though this is not a perfect oracle because Atari actions are context-dependent and can't simply be treated as one abstract action group, we can still use this approach to characterize the action-generalization gap.

Figure~\ref{fig:large-scale} (right) shows a large action-generalization gap for the duplicate  and full action set augmentations, across the majority of games. We hypothesize that this degradation occurs because 1) Atari games are more complicated domains and 2) Atari games have a larger action space to begin with, both of which make it harder for DQN to generalize over actions. Surprisingly, the noop augmentation only leads to an action-generalization gap on Ms Pacman. We suspect that which specific actions lead to worse performance may be game-dependent, and that in the other games, noop actions may more frequently be the optimal choice.

\section*{Conclusion}
Overall, we have obtained preliminary evidence suggesting that DQN has some ability to do action generalization in small to medium action spaces, but for the most part that ability does not extend to large action spaces. However, given enough time, DQN may still be able to recover the optimal policy in those large spaces. A direction for future work is to pinpoint the exact reason for DQN not being able to generalize over actions in large action spaces and to provide a remedy that improves performance.

\bibliographystyle{apalike}
\bibliography{rldm}

\end{document}